\documentclass[sigconf]{acmart}

\usepackage{booktabs} 
\usepackage{todonotes}
\usepackage{tabu}
\usepackage[ruled]{algorithm2e}
\usepackage{soul}
\usepackage{algorithmic}

\usepackage{color, colortbl}
\definecolor{Gray}{gray}{0.9}
\definecolor{light-gray}{gray}{0.95}

\usepackage{hyperref}

\begin{document}
\title{Comparing and Combining Lexicase Selection and Novelty Search}

\author{Lia Jundt}
\affiliation{%
  \institution{Hamilton College}
  \streetaddress{198 College Hill Road}
  \city{Clinton} 
  \state{New York} 
  \country{USA}
}
\email{liajundt@gmail.com}

\author{Thomas Helmuth}
\orcid{0000-0002-2330-6809}
\affiliation{%
  \institution{Hamilton College}
  \streetaddress{198 College Hill Road}
  \city{Clinton} 
  \state{New York} 
  \country{USA}
}
\email{thelmuth@hamilton.edu}


\begin{abstract}
Lexicase selection and novelty search, two parent selection methods used in evolutionary computation, emphasize exploring widely in the search space more than traditional methods such as tournament selection. However, lexicase selection is not explicitly driven to select for novelty in the population, and novelty search suffers from lack of direction toward a goal, especially in unconstrained, highly-dimensional spaces. We combine the strengths of lexicase selection and novelty search by creating a novelty score for each test case, and adding those novelty scores to the normal error values used in lexicase selection. We use this new novelty-lexicase selection to solve automatic program synthesis problems, and find it significantly outperforms both novelty search and lexicase selection. Additionally, we find that novelty search has very little success in the problem domain of program synthesis. We explore the effects of each of these methods on population diversity and long-term problem solving performance, and give evidence to support the hypothesis that novelty-lexicase selection resists converging to local optima better than lexicase selection.
\end{abstract}

%
%

\copyrightyear{2019} 
\acmYear{2019} 
\setcopyright{acmcopyright}
\acmConference[GECCO '19]{Genetic and Evolutionary Computation Conference}{July 13--17, 2019}{Prague, Czech Republic}
\acmBooktitle{Genetic and Evolutionary Computation Conference (GECCO '19), July 13--17, 2019, Prague, Czech Republic}
\acmPrice{15.00}
\acmDOI{10.1145/3321707.3321787}
\acmISBN{978-1-4503-6111-8/19/07}



\maketitle

\section{Introduction}

In many problems tackled by genetic programming (GP), each individual is evaluated on a number of different \textit{test cases} to determine how well it solves the problem. In order to determine which individuals perform better and should be selected as parents, we calculate a \textit{fitness error} for each individual on each test case by comparing the distance between each output of the program and the correct output.
Most parent selection methods, including tournament selection, aggregate all of the errors of an individual into a single \textit{fitness value}, which is used to determine which individuals to select as parents. 

Lexicase selection is a parent selection algorithm that does not combine errors into a single fitness value, but instead treats them all separately~\cite{LexicaseUncomprimising, LexicaseOriginal}. Lexicase selection often selects \textit{specialists}, or individuals that perform extremely well on one or several test cases, but may not perform well on others, and thus may have a poor aggregate fitness value~\cite{Helmuth:2019:GECCO:Specialists}. Specialists are usually good at solving test cases with similar behavior, but may not have figured out how to solve test cases with a different pattern. Lexicase selection, by favoring specialists, allows the GP system to often explore many areas of the search space simultaneously instead of converging on one promising-looking area of behavior. Lexicase selection has been shown to increase the diversity of the population, and improves performance on several types of problems, including software synthesis problems~\cite{LexicaseDiversity, lexicaseVtourn}.

With a similar motivation of avoiding premature convergence of evolutionary computation to areas of the search space with no solutions, Lehman and Stanley proposed novelty search~\cite{Lehmanoriginal, Lehman_Efficiently}.
Novelty search aims to explore the search space by selecting individuals with unique or uncommon behavior to be parents, without considering the fitness errors of the potential parents. In this way, novelty search does not drive selection towards an objective, but instead towards individuals that are different from those seen previously.
Novelty search has proven itself as an excellent selection method for solving problems with deceptive fitness landscapes~\cite{Lehmanoriginal, Lehman_Efficiently, Lehman11, Revising, Empirical_Study}, but has not yet been tested on general program synthesis problems.


In this paper, we synthesize novelty search and lexicase selection into \textit{novelty-lexicase} selection, with the aim of exploring and exploiting the search space by encouraging novel and specialized behavior simultaneously. We combine these objectives by adding ``novelty scores'' to the normal error values used by lexicase selection. Novelty-lexicase selection at times emphasizes selection based on novelty; other times it emphasizes good performance on the test cases; often it selects a single individual based on some aspects of both.


We conducted experiments comparing novelty-lexicase selection with novelty search, lexicase selection, and tournament selection on a suite of general program synthesis benchmark problems, a new problem domain for novelty search. While experiments capped at 300 generations failed to differentiate novelty-lexicase and lexicase selections, other metrics suggested that longer runs may benefit novelty-lexicase. Indeed, on runs capped at 1000 generations, novelty-lexicase selection significantly outperformed lexicase selection. This indicates that although novelty-lexicase selection is slower to find solutions on average, it continues to effectively search for solutions longer than standard lexicase selection.

The following section gives background information, and we give the details of novelty-lexicase selection in section~\ref{sec:novlex}. We next explain our experiments with novelty-lexicase selection in section~\ref{sec:experiments}, give the results of those experiments in section~\ref{sec:results}, and discuss the results in section~\ref{sec:discussion}.

\section{Related Work}
We will discuss behavioral diversity, a method for measuring population diversity, and the two selection methods motivating novelty-lexicase selection.

\subsection{Behaviors and Diversity}

We define the \textit{behavior} of an individual to be the vector of its outputs over the inputs. 
Behavioral diversity is defined as the proportion of the population with distinct behaviors~\cite{JacksonPaper}. In order to calculate this, we tally the number of distinct behaviors within the population, and then divide by the number of individuals. Behavioral diversity is a phenotypic measure of diversity---meaning it is a diversity measure based on the functionality of programs, rather than their code or structure. An increased behavioral diversity allows the GP system to search programs in various parts of the search space simultaneously, and explore multiple methods of solving the problem. Previous studies have shown behavioral diversity to correlate well with problem-solving performance~\cite{JacksonPaper}.

\subsection{Lexicase Selection} \label{sec:lexicase}

\begin{algorithm}[t]
\caption{Lexicase Selection \textit{(to select one parent)}}
\label{alg:lexicase}
\begin{algorithmic}
\STATE Inputs: $candidates$, the entire population;
\STATE $\phantom{Inputs: } cases$, a list of test cases
\STATE Shuffle $cases$ into a random order
\LOOP
	\STATE Set $first$ be the first case in $cases$
	\STATE Set $best$ be the best performance of any individual in $candidates$ on the $first$ test case
	\STATE Set $candidates$ to be the subset of 
	$candidates$ that have exactly $best$ performance on $first$
	\IF {$|candidates| = 1$}
	    \STATE Return the only individual in $candidates$
	\ENDIF
	\IF {$|cases| = 1$}
	    \STATE Return a randomly selected ind. from $candidates$
	\ENDIF
	\STATE Remove the first case from $cases$
\ENDLOOP
\end{algorithmic}
\end{algorithm}

Lexicase selection can be used in any setting where individuals are evaluated on multiple test cases; in our study, test cases are given as input/output pairs, as in supervised learning. Lexicase selection chooses a parent by pruning the population over a number of steps. During each step, it compares the remaining candidate pool on a single randomly-selected test case and removes any individuals that do not have the very best error on this test case. This process continues until one individual remains or all test cases have been used, at which point it returns a randomly selected candidate from those remaining. Algorithm~\ref{alg:lexicase} formalizes lexicase selection.
While lexicase selection has typically been used over the errors of the individuals on each test case, it could be used on any type of scores assigned to the members of the population.

Previous studies have shown lexicase selection outperforming tournament selection and other parent selection methods in areas such as automatic program synthesis~\cite{ProblemInfo, Forstenlechner:2017:EuroGP}, boolean logic and finite algebras~\cite{LexicaseUncomprimising, Helmuth:2013:GECCOcomp, Liskowski:2015:GECCOcomp}, evolutionary robotics~\cite{moore:2017:ecal}, and boolean constraint satisfaction using genetic algorithms~\cite{Metevier2019}. Additionally, $\epsilon$-lexicase, a relaxed version of lexicase selection that at each step keeps any individuals within some threshold of the best individual, has performed well on symbolic regression problems~\cite{LaCava:2018:ecj, LaCava:2016:GECCO}. Studies showing that lexicase selection produces higher population diversity than other methods help explain its improved performance; it explores more widely in the search space, not converging to local optima, while still applying pressure for improvement on the test cases~\cite{LexicaseDiversity, lexicaseVtourn}.

\subsection{Novelty Search}

In 2008, Lehman and Stanley first presented novelty search as a way to avoid convergence to local optima when searching for a solution to a problem with a deceptive fitness landscape~\cite{Lehmanoriginal}. Fitness-seeking algorithms tend towards convergence, narrowing their search towards the most promising part of the search space. In many problems though, such as maze solving, converging towards the seemingly best solution can often lead to local optima where there is in fact no solution to be found. Novelty search originated from a need to prevent this convergence to local optima and increase genetic diversity within the population~\cite{Lehmanoriginal, Lehman_Efficiently, Lehman11}. Instead of selecting individuals with good performance, novelty search selects the most novel individuals to be parents for the next generation.

Novelty is calculated as the distance between an individual's behavior vector and that of its $k$ closest neighbors in the population~\cite{Lehman_Efficiently, Empirical_Study}. Some studies maintain a novelty archive consisting of individuals from earlier populations; these archives are also considered when finding the $k$ closest neighbors~\cite{Lehmanoriginal, Empirical_Study}. Once the novelty scores have been calculated, novelty search selects parents with tournament selection, selecting the most novel individual in each tournament regardless of their fitness. Novelty search has typically been applied to problems where behaviors have a small number of dimensions, allowing for inexpensive distance calculations compared to problems with behaviors over many dimensions.

Novelty search has proven useful in a variety of fields, such as Grammatical Evolution \cite{GEinSantaFeTrail}, Soft Robotics \cite{LaCava:2018:ALIFE} and Swarm Robotics \cite{Gomes2013}, though in soft robotics and swarm robotics it did not significantly improve results. Additionally, novelty search has been proven effective for solving problems in GP, such as difficult symbolic regression problems \cite{Martinez:2013:CEC}.
In previous studies, novelty search has proven more effective than fitness search for problems with deceptive fitness landscapes, such as the maze and artificial ant problems\cite{Lehman_Efficiently, GEinSantaFeTrail}. 

Additionally, a family of related search techniques, termed ``quality diversity'' algorithms, aim to combine the ideals of novelty search with performance optimization~\cite{QualityDiversity}. Quality diversity algorithms are motivated by the same issues with novelty search that motivate us here to combine novelty search's ability to explore the search space while also providing pressure toward solving the problem. These quality diversity algorithms include MAP-elites~\cite{DBLP:journals/corr/MouretC15}, using a weighted sum of an individual's novelty and fitness scores, multiobjectivisation of novelty and fitness scores, and progressive minimal criteria novelty search (PMCNS) where individual's novelty scores are penalized for being below the 50th percentile in fitness score~\cite{Empirical_Study}.  MAP-Elites aims to optimize performance within bins that emphasize maintaining diversity through novelty in the population~\cite{DBLP:journals/corr/MouretC15, Dolson}. Multiobjectivisation of novelty and fitness scores and the weighted score methods worked well on problems with deceptive fitness landscapes, but the PMCNS method did not~\cite{Empirical_Study}.

\subsection{Combining Novelty and Lexicase}
 
Perhaps the most similar work to that discussed here is another effort to combine novelty search and objective performance, called knobelty~\cite{kelly2019}. This work takes a different approach, with each parent selection event choosing to either select based on novelty (using tournament selection) or based on performance (using lexicase selection). Thus the novelty-based selection does not use the lexicase algorithm. They showed that knobelty, when paired with grammatical evolution and a variety of novelty metrics, performed well compared to vanilla lexicase selection on three program synthesis benchmark problems.





\section{Novelty-Lexicase Selection} \label{sec:novlex}

Novelty-lexicase selection uses the standard lexicase selection algorithm presented in section~\ref{sec:lexicase}, except that we add novelty scores to the list of error values normally used by lexicase selection. Similarly to how lexicase selection typically uses one error value per test case, we create one novelty score per test case. Thus, the list of cases used by novelty-lexicase selection is a randomly shuffled combination of errors and novelty scores, and each step of the algorithm compares the population either on one error value or one novelty score. Since lexicase selection puts the most emphasis on performance of the first handful of cases in the shuffled case list, some uses of novelty-lexicase will randomly emphasize excellence on novelty scores, some will emphasize excellence on fitness errors, and some will emphasize a mix of both. In this way, novelty-lexicase selection drives the population simultaneously towards novel behavior and more fit behavior. 

Each novelty score should measure how similar a program's behavior is to other behaviors in the population.
In previous novelty search studies, the novelty of an individual is calculated as the average distance to the nearest $k$ neighbors, where the distance is measured between the individuals' behavior vectors~\cite{Empirical_Study,Enhancements,Lehman_Efficiently,Revising}. In order to use lexicase, we would instead like to calculate novelty scores for each test case. As computing the average distance to the $k$ closest neighbors over hundreds of test cases for each individual each generation incurs considerable expense when comparing outputs such as strings, we instead use a simpler novelty metric. We define the \textit{novelty score} of one behavior (output) of an individual to be the number of individuals in the population and archive that produce the same behavior on that test case. A lower novelty score indicates increased novelty, since fewer individuals produce the same output---thus novelty-lexicase selection will prefer lower novelty scores, as it prefers lower errors.

As recommended with novelty search, novelty-lexicase selection maintains an archive of individuals, adding one randomly selected individual each generation. We calculate novelty scores with respect to the combined population and archive, encouraging individuals to be novel not only with respect to their population, but also to areas of the search space explored in prior generations.

Novelty-lexicase selection never combines novelty scores into a single aggregate value, distinguishing it from novelty search and other quality diversity algorithms. As with lexicase selection not combining errors into a single fitness value, this allows novelty-lexicase to differentiate between the performance/novelty of different outputs. Thus an individual that exhibits extremely novel behavior on some test cases but very common behavior on other test cases may receive the attention of novelty-lexicase selection, where it would have received little or no attention by novelty search. This allows novelty-lexicase to select individuals that specialize in not only excellent performance on some test cases, but also excellent novelty on some test cases.

There is no asymptotic difference in time complexity between novelty-lexicase and lexicase selections, as they simply follow the same algorithm, except that novelty-lexicase has twice as many test cases in its list of test cases. The time for calculating novelty scores is negligible compared to the run time of the lexicase algorithm.


\section{Experimental Design} \label{sec:experiments}

Below we discuss the benchmark problems, GP system, and GP parameters used in our experiments.

\subsection{Benchmark Problems}

\begin{table}
\centering
\caption{Benchmark problem input and output types. CSL is the Compare String Lengths problem, and RSWN is the Replace Space With Newline problem.}
\label{table:Problems}
\taburowcolors[2] 2{white .. light-gray}
\begin{tabu}{l ll}
\toprule
\textbf{Problem} & \textbf{Input Type} & \textbf{Output Type}\tabularnewline
\midrule
CSL & 3 Strings & Boolean \tabularnewline
Double Letters & String & Printed String \tabularnewline
Last Index of Zero & Vector of Integers & Integer \tabularnewline
Mirror Image & 2 Vectors & Boolean \tabularnewline
Negative to Zero & Vector of Integers & Vector of Integers \tabularnewline
RSWN & String & Printed String, Int \tabularnewline
Scrabble Score & String & Integer \tabularnewline
Syllables & String & Printed String \tabularnewline
Vector Average & Vector of Floats & Float \tabularnewline
X-Word Lines & String, Integer & Printed String \tabularnewline
\bottomrule
\end{tabu}
\end{table}

We compare selection strategies on ten representative program synthesis benchmark problems~\cite{ProblemInfo}. These problems challenge the GP system to produce programs that mimic introductory computer science student code, requiring various control flow structures and data types. We chose this subset of problems from the benchmark suite as they represent a wide range of difficulties, with some rarely solved in previous studies while others have been solved more than 70\% of the time using standard lexicase selection. Additionally, we chose problems with diverse data type requirements; we list the input and output types for each problem in Table~\ref{table:Problems}. We decided to use program synthesis benchmark problems instead of problems with deceptive fitness landscapes (such as the maze problems that have frequently been used with novelty search~\cite{Lehman_Efficiently, MinCriterion, QualityDiversity, illuminationalgos, Empirical_Study}) to highlight the differences between novelty search and novelty-lexicase selection on problems with highly-dimensional behaviors---the problems here each have 100 or more test cases used for evaluation.

For novelty search, we need to define the distance metrics used for the different possible output types listed in Table~\ref{table:Problems}.
The problems Compare String Lengths (CSL) and Mirror Image produce boolean outputs, meaning distances between behavior vectors are comparing vectors of booleans, for which it is natural to use Hamming distance. For the string and integer outputs of Replace Space With Newline (RSWN), we used Hamming distance simply comparing whether each output was equal or not to the other individual's output. The Last Index of Zero and Vector Average problems output integers/floats respectively, allowing us to use the Manhattan distance between vectors of numbers. Manhattan distance gives a more accurate estimate of the distance between two output vectors. However, we found Manhattan distance to be prohibitively expensive to calculate on problems with string output types, since it requires a Levenshtein distance calculation on every pair of string outputs for the same test case in the population. For this reason, we use a simple Hamming distance comparison for string outputs. For the Negative to Zero problem, which outputs a vector of integers for each test case, it was unclear to us how one would best calculate distances between vectors of vectors of integers.

\subsection{GP System and Parameters}

Our experiments with novelty-lexicase selection use PushGP, a GP system that evolves programs in the Push programming language~\cite{1068292, spector:2002:GPEM}.
Push maintains a stack for each data type, taking arguments from the stacks and pushing return values to the stacks. Additionally, it allows for a variety of standard and exotic control flow structures, aided by the fact that the running Push program itself is stored on a stack.
Push was specifically designed for use in GP, contributing to its flexibility with data types and its simple syntax, in which every nested list of instructions and literals is a valid Push program.
We use the Clojure\footnote{\url{https://github.com/lspector/Clojush}} implementation of PushGP, which has previously been used in a number of program synthesis experiments~\cite{ProblemInfo, LexicaseDiversity, lexicaseVtourn}. The source code used in our experiments is freely available.\footnote{\url{https://github.com/thelmuth/Clojush/releases/tag/Novelty-Lexicase}}

To measure the performance of each selection method, we follow the recommendation of the benchmark suite to measure the number of successful runs out of 100 GP runs~\cite{ProblemInfo}. We stop a GP run if it finds a program that perfectly passes all of the test cases used as training data. Additionally, for a program to count as a \textit{generalizing} success, it must perfectly pass a large number of withheld test cases not used during evolution. Before testing for generalization, we automatically simplify each solution program, which has been shown to significantly raise generalization rates~\cite{Helmuth:2017:GECCO}. To test for significant differences in success rates, we use a pairwise chi-square test with Holm correction and a 0.05 significance level.

\begin{table}
\centering
\caption{GP system parameters and the usage rates of genetic operators.}
\label{table:PushGP-params}
\begin{tabular}{l r}
\toprule
\textbf{Parameter} & \textbf{Value} \tabularnewline
\midrule
population size & 1000 \tabularnewline
max number of generations & 300, 1000 \tabularnewline
tournament size for tournament selection & 7 \tabularnewline
tournament size for novelty search & 2 \tabularnewline
individuals added to archive per generation & 1 \tabularnewline
$k$ nearest neighbors for novelty search & 25 \tabularnewline
\midrule
\textbf{Genetic Operator Rates} & \textbf{Prob} \tabularnewline
\midrule
alternation & 0.2 \tabularnewline
uniform mutation & 0.2 \tabularnewline
uniform close mutation & 0.1 \tabularnewline
alternation followed by uniform mutation & 0.5 \tabularnewline
\bottomrule
\end{tabular}
\end{table}

We conducted 100 runs in each configuration on each benchmark problem using the GP parameters in Table~\ref{table:PushGP-params}. Except for where stated otherwise, we use the same parameters and recommendations given in the description of the program synthesis benchmark suite~\cite{ProblemInfo}. While we initially ran each run for a maximum of 300 generations, the results suggested that longer evolution times might prove interesting (as we will discuss below), so we also conducted experiments using a maximum of 1000 generations. For novelty search and novelty-lexicase selection, we add one randomly selected individual to the archive each generation, as recommended by Gomes et al.~\cite{Empirical_Study}. For novelty search, we assign novelty as the average distance between an individual and its $k = 25$ nearest neighbors, use a tournament size of 2, and a population of 1000, as recommended by Lehman and Stanley~\cite{Lehman_Efficiently}.

\section{Results} \label{sec:results}


\begin{table}
    \centering
    \caption{Number of successes out of 100 runs for novelty-lexicase selection, lexicase selection, tournament selection, and novelty search with a maximum of 300 generations. We have not run novelty search on Negative to Zero. Bold values are significantly better than all others in that row.}
    \label{tab:completionrates}
    \taburowcolors[2] 2{white .. light-gray}
	\begin{tabu} {l |rrrr}
		\toprule
        \textbf{Problem} & \textbf{Novelty-lex} & \textbf{Lex} & \textbf{Tourn} & \textbf{Novelty}\\
        \midrule
        CSL & 3 & 5 & 3 & 0 \\
        Double Letters & 4 & 1 & 0 & 0 \\
        Last Index of Zero & 35 & 29 & 4 & 1 \\
        Mirror Image & \textbf{100} & 87 & 45 & 72 \\
        Negative to Zero & 61 & 62 & 6 & - \\
        RSWN & 71 & 58 & 15 & 0 \\
        Scrabble Score & 1 & 4 & 0 & 0 \\
        Syllables & 11 & 24 & 2 & 0 \\
        Vector Average & 35 & 43 & 14 & 0 \\
        X-Word Lines & 18 & 18 & 0 & 0 \\
        \bottomrule
    \end{tabu}
\end{table}

Table~\ref{tab:completionrates} gives the number of successes out of 100 runs on each problem for each selection method with a maximum of 300 generations. Both lexicase selection and novelty-lexicase selection outperformed tournament selection and novelty search consistently. The lexicase selection and novelty-lexicase selection results are very similar, with the only significant difference between them on the Mirror Image problem. Novelty search performed particularly poorly here, only finding solutions to two of the nine problems on which we ran it, and only outperforming tournament selection on one problem.


\begin{figure}
    \centering
    \includegraphics[width=0.99\linewidth]{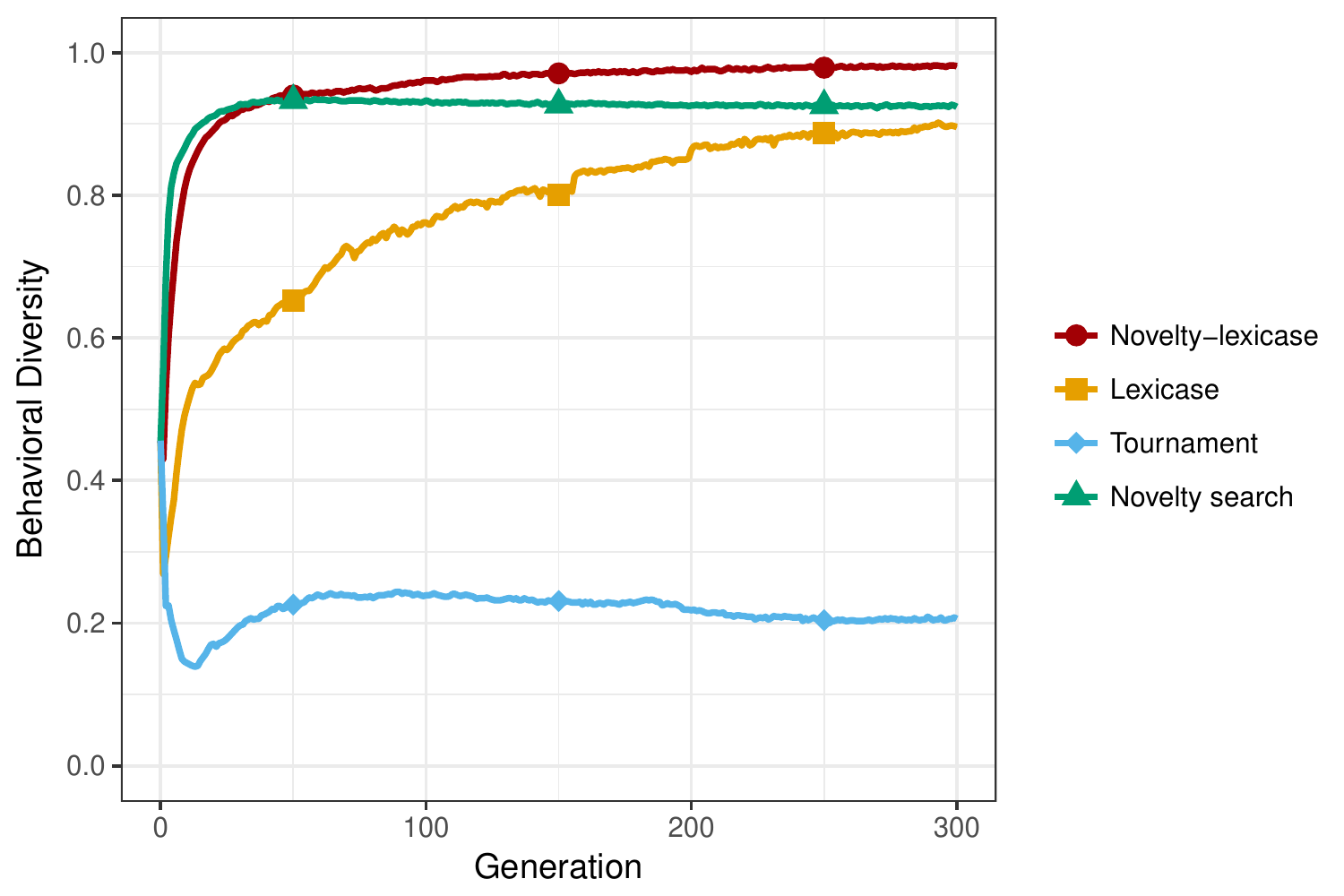}
    \caption{Mean population behavioral diversity across generations of each selection method on the Replace Space with Newline problem.}
    \label{fig:bdrswn}
\end{figure}

\begin{figure}
    \includegraphics[width=0.99\linewidth]{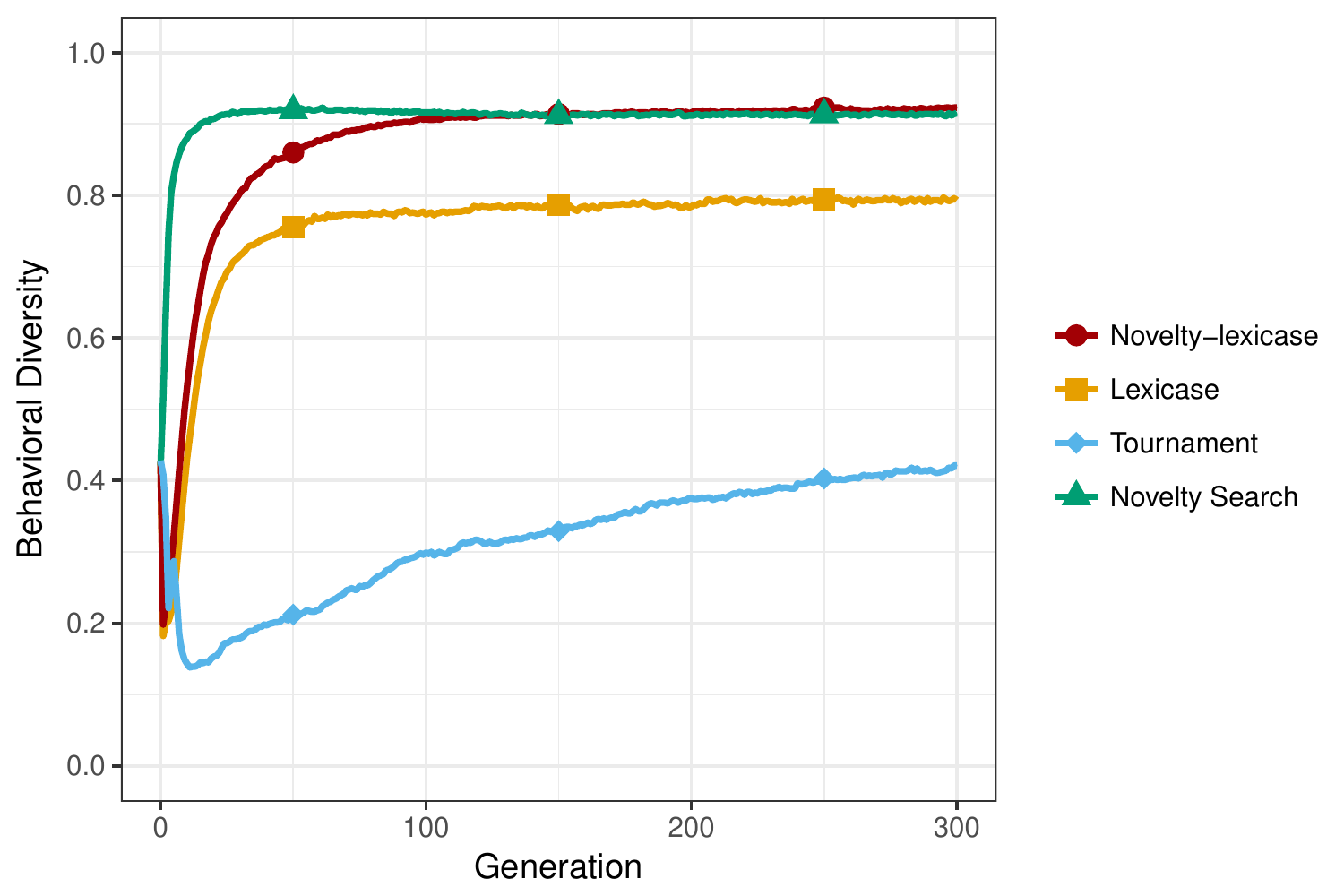}
    \caption{Mean population behavioral diversity across generations of each selection method on the Syllables problem.}
    \label{fig:bdsyllables}
\end{figure}

Figures \ref{fig:bdrswn} and \ref{fig:bdsyllables} show the behavioral diversity of each selection method over the maximum 300 generations on two representative problems, Replace Space with Newline and Syllables.\footnote{The initial dip in behavioral diversity in the first few generations is a known phenomenon, caused by a few randomly generated individuals dominating the selection, and should be ignored in favor of the later trends~\cite{Helmuth:2016:GECCO}.} The levels of diversity in these plots mirror those for other problems, which we do not present. Although novelty-lexicase selection did not achieve a significantly better success rate than lexicase selection, the behavioral diversity of the novelty-lexicase populations was close to 1 for both problems, much higher than lexicase selection, which in turn is much higher than tournament selection. Both novelty search and novelty-lexicase selection rapidly evolve diverse populations, while lexicase selection cultivates diversity more slowly on Replace Space with Newline. When the diversity of novelty search peaks and levels off around generation 30, the diversity of novelty-lexicase continues to rise throughout evolution. As has been noted previously~\cite{lexicaseVtourn}, tournament selection does an extremely poor job of maintaining population diversity compared to the other methods.


Even when solving the Syllables problem, where novelty-lexicase selection produced fewer solutions than lexicase selection, novelty-lexicase selection created more behaviorally diverse populations than lexicase or tournament selection. In this case, lexicase and novelty-lexicase selection's behavioral diversity grew significantly in the first twenty generations, but then lexicase plateaued just under 0.80 behavioral diversity, while novelty-lexicase hovered over 0.90. Novelty search grows diversity quicker than any other method, but is again overtaken slightly by novelty-lexicase selection my the end of the runs. Tournament selection's behavioral diversity was, at it's highest point near the end of the run, approximately 0.40.

\begin{table}
    \centering
    \caption{Average generation of solution discovery for novelty-lexicase selection, lexicase selection, tournament selection, and novelty search with a maximum of 300 generations.}
    \label{tab:gensolution}
    \taburowcolors[2] 2{white .. light-gray}
	\begin{tabu} {l | cccc}
		\toprule
        \textbf{Problem} & \textbf{Novelty-lex} & \textbf{Lex} & \textbf{Tourn} & \textbf{Novelty} \\
        \midrule
        CSL & 199 & 13 & 28 & - \\
        Double Letters & 270 & 203 & - & - \\
        Last Index of Zero & 83 & 81 & 15 & 250 \\
        Mirror Image & 44 & 18 & 43 & 100 \\
        Negative to Zero & 85 & 80 & 141 & - \\
        Scrabble Score & 239 & 128 & - & - \\
        Syllables & 237 & 159 & 84 & - \\
        RSWN & 96 & 93 & 122 & - \\
        Vector Average & 150 & 138 & 66 & - \\
        X-Word Lines & 239 & 199 & - & - \\
        \bottomrule
    \end{tabu}
\end{table}

The behavioral diversity of the novelty-lexicase populations were consistently higher than those produced by tournament selection or lexicase selection; higher than lexicase selection by more than 5\% consistently. This higher diversity raises the question of which algorithm is finding solutions faster; whether the diversity of novelty-lexicase selection allows it to locate important areas of the search space first, or if it is putting more effort into exploration and less into exploitation of good programs, leading to slower traversal of gradients toward solutions. We calculated the average generation of solution discovery in Table~\ref{tab:gensolution}. Novelty-lexicase selection was generally slower than lexicase and tournament selections, only twice finding solutions faster than tournament selection. Lexicase selection was consistently faster at finding solutions than novelty-lexicase selection, although on the Last Index of Zero, Negative to Zero, and Replace Space With Newline problems the difference was small.

Since we see that novelty-lexicase selection produces increased diversity and takes more generations to find solutions, we hypothesize that with a larger number of generations novelty-lexicase will continue to find solutions, while lexicase selection will slow down as remaining runs sometimes converge to local optima. Thus we conducted a second experiment, where novelty-lexicase and lexicase systems run for 1000 generations.

\subsection{Maximum of 1000 Generations}


Table~\ref{tab:completionrates1000} presents the results of extended, 1000-generation runs. While both novelty-lexicase and lexicase selection increased their solution rates on most problems compared to 300 generations, many of lexicase's increases were insignificant while novelty-lexicase produced large increases on most problems. Novelty-lexicase selection performed significantly better than lexicase selection on five problems, while lexicase only performed significantly better on one.

\begin{figure}
    \includegraphics[width=0.99\linewidth]{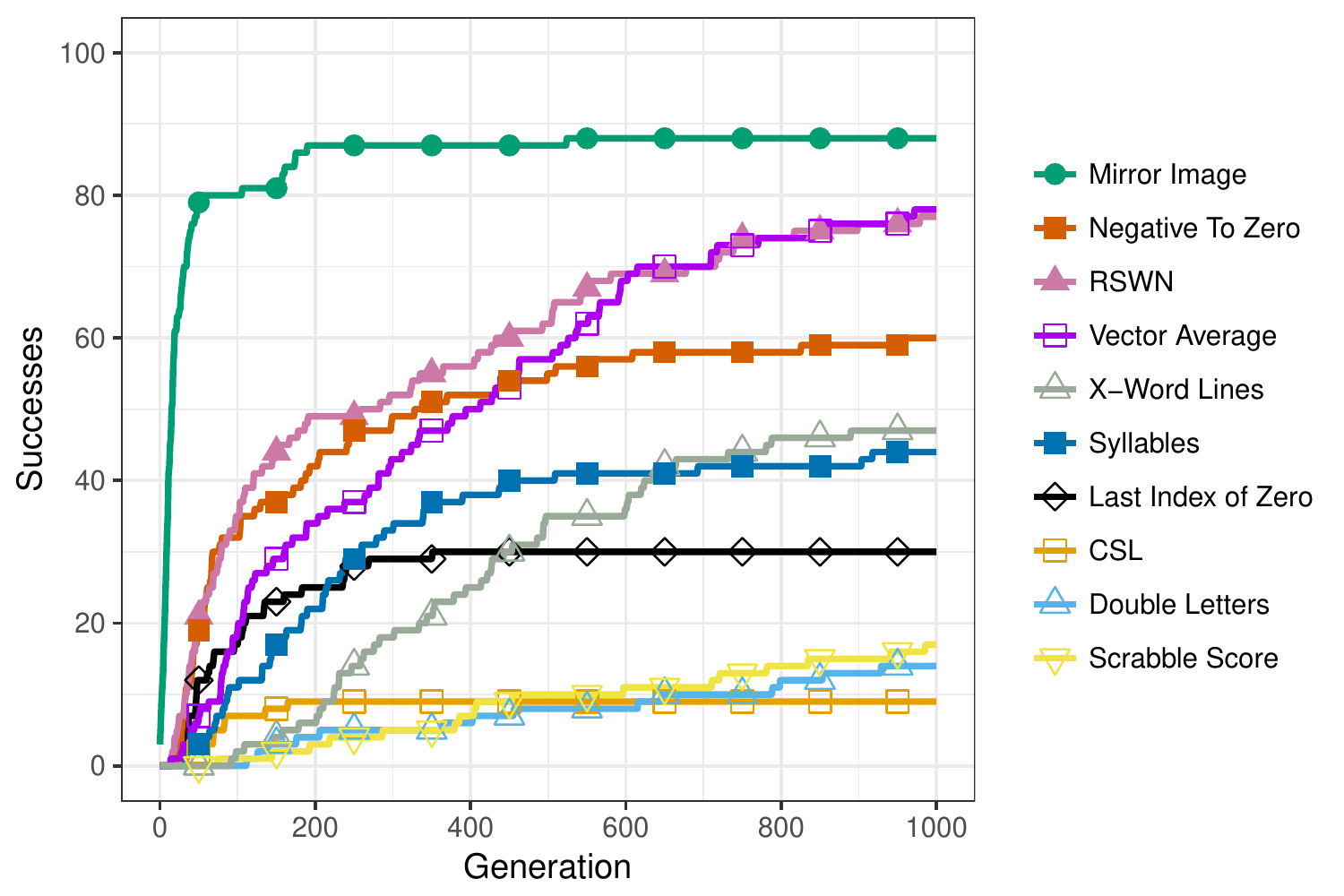}
    \caption{The accumulated number of solutions found by lexicase selection over 1000 generations for every problem.}
    \label{fig:lexicase-gen}
\end{figure}

\begin{table}
    \centering
    \caption{Number of successes out of 100 runs for novelty-lexicase selection and lexicase selection with a maximum of 1000 generations. Results that are significantly better at the 0.05 level are in bold.}
    \label{tab:completionrates1000}
    \taburowcolors[2] 2{white .. light-gray}
	\begin{tabu} {l |cc}
		\toprule
        \textbf{Problem} & \textbf{Novelty-lex} & \textbf{Lex}\\
        \midrule
        Compare String Lengths & \textbf{39} & 9  \\
        Double Letters & 17 & 14  \\
        Last Index of Zero & \textbf{51} & 30 \\
        Mirror Image & \textbf{100} & 88 \\
        Negative to Zero & \textbf{83} & 60 \\
        RSWN & 77 & 77 \\
        Scrabble Score & 18 & 17 \\
        Syllables & 50 & 44 \\
        Vector Average & 62 & \textbf{78} \\
        X-Word Lines & \textbf{65} & 47 \\
        \bottomrule
    \end{tabu}
\end{table}

\begin{figure}
    \includegraphics[width=0.99\linewidth]{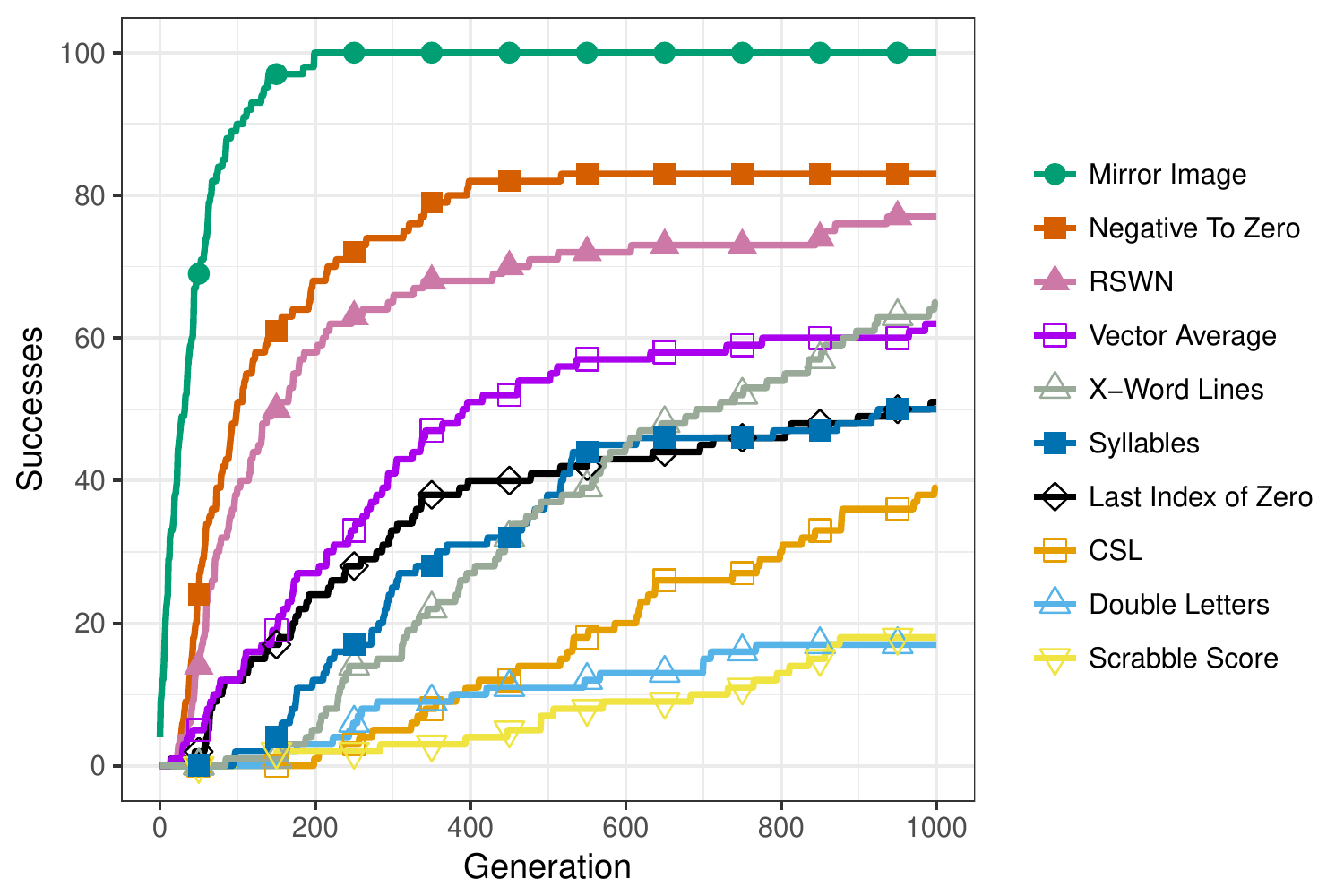}
    \caption{The accumulated number of solutions found by novelty-lexicase selection over 1000 generations for every problem.}
    \label{fig:nov-lex-gen}
\end{figure}

\begin{figure}
    \includegraphics[width=0.99\linewidth]{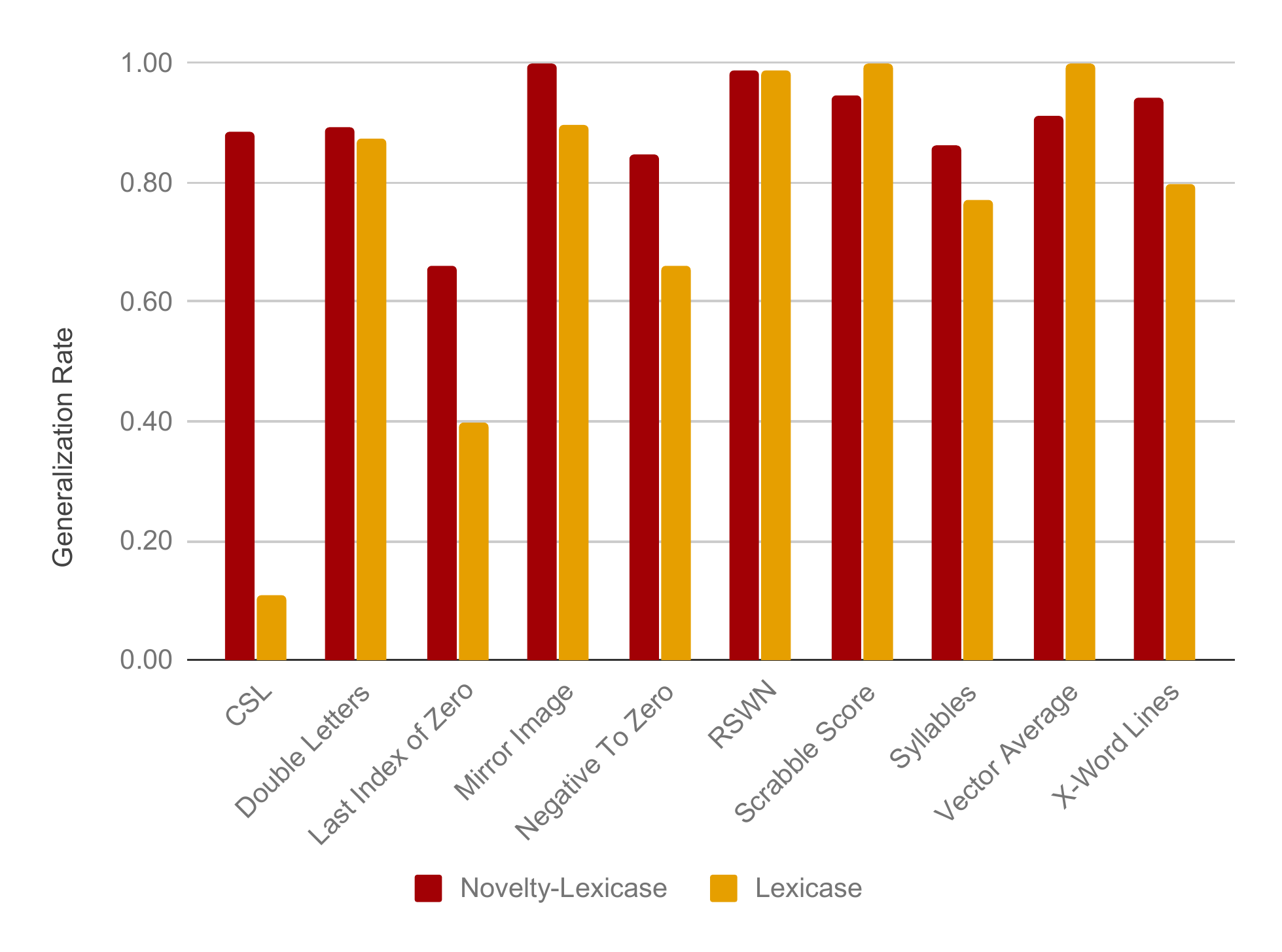}
    \caption{Generalization rates of novelty-lexicase and lexicase selections (after simplification).}
    \label{fig:generalization}
\end{figure}

To get a better understanding of when during search GP finds solutions, we plot the accumulated number of solutions found over all 1000 generations in Figures~\ref{fig:lexicase-gen} and \ref{fig:nov-lex-gen}. Figure~\ref{fig:lexicase-gen} shows that lexicase selection finds many of its solutions in the first 500 generations. On the other hand, novelty-lexicase continues finding solutions to most problems right up to generation 1000, and may even continue finding solutions past generation 1000 if it were given the chance. Take, for example, the Compare String Lengths (CSL) problem. As can be seen in Figure \ref{fig:lexicase-gen}, the number of successes for lexicase selection plateaus around generation 200, while the number of successes found by novelty-lexicase selection in Figure \ref{fig:nov-lex-gen} steadily increases over all 1000 generations. 

In general, lexicase selection plateaus and stops finding new solutions around generation 500, while novelty-lexicase selection does not have a clear plateau, except for the Mirror Image and Negative To Zero problems. The plateau in solutions found for the Mirror Image problem is due to all runs having found solutions by this point. The Negative To Zero problem plateaus for a similar reason: 98 out of 100 runs found a solution that passed the training data, and hence stopped evolution. However, only about 80\% of these solutions generalized and counted as a success. So although this set of runs plateaus at 83 successes, only 2 runs continued beyond generation 600 and could have contributed to more successes.

Although novelty-lexicase selection had some difficulty on that problem with generalization, overall novelty-lexicase had better generalization of training solutions to unseen test data than lexicase selection, as shown in Figure~\ref{fig:generalization}. The difference in generalization rate is significant in the favor of novelty-lexicase selection on the Compare String Lengths, Last Index of Zero, Mirror Image, Negative to Zero, and X-Word Lines problems, while lexicase selection has significantly better generalization on only the Vector Average problem.  For whatever reason, lexicase selection finds many more solutions to the training data that do not generalize to unseen data. While we have not explored exactly what causes this difference, we offer a hypothesis in the next section. Regardless of why novelty-lexicase generalizes better, its ability to generalize contributes significantly to its better success rate on many of these problems.

\section{Discussion} \label{sec:discussion}

When comparing the diversity rates produced by each selection method, we found that novelty search and novelty-lexicase have many similarities, with both displaying higher levels of diversity than lexicase selection. By shuffling the errors normally used by lexicase selection with novelty scores on each test case, we have added extra objectives that focus on individuals that not only perform well, but behave differently from other individuals in the population, leading to increased diversity.

Is GP exploring additional interesting and useful parts of the search space by creating more novel individuals, or simply increasing novelty for novelty's sake while using up program evaluations that could instead be used for refining promising individuals? The low success rates produced by novelty search indicate that in the highly-dimensional and unbounded output spaces of the program synthesis problems considered here, novelty search often produces a huge variety of individuals without finding a solution. Novelty search does not apply selection pressure toward better individuals by design; while this enables it to produce interesting and different individuals, as well as solve problems in constrained, low-dimensional output spaces, there are simply too many ways to be novel without being good when a behavior contains 100 string or integer outputs. We note that novelty search may perform better on these problems with a more carefully crafted method for measuring behavior than simply listing the outputs of the program on the test cases, but with no obvious alternative, this nontrivial task is beyond the scope of this paper.

Turning to novelty-lexicase selection, when running GP for a maximum of 300 generations, we found that the increased diversity did not immediately translate into more successes than lexicase selection, though they behaved comparably. However, given more time to evolve, novelty-lexicase selection converged to local optima much less often than lexicase selection, finding solutions throughout all 1000 generations while lexicase selection often failed to find many solutions after 300-500 generations. The lexicase results mirror some extended 600 generation runs on this same benchmark suite using grammar-guided GP, where lexicase selection often plateaued after approximately 300 generations~\cite{Forstenlechner:2018:CEC}. The ability of novelty-lexicase selection to avoid premature convergence and continue effectively exploring the search space shows great promise whenever evolution requires many generations to construct complex solution programs.

In addition to avoiding local optima, novelty-lexicase selection tends to evolve solutions that more likely generalize than those produced by lexicase selection, as shown in Figure~\ref{fig:generalization}. We hypothesize that the strong selection pressure that lexicase selection applies toward finding solutions encourages solutions that sometimes ``memorize'' edge cases in a way that does not generalize. Additionally, because novelty-lexicase selection gives the lexicase algorithm a combination of novelty scores and fitness errors, any one fitness error will appear in the first $n$ shuffled cases half as often as with standard lexicase selection. Thus novelty-lexicase selection will reward memorizing the answer to a single edge case less often than lexicase selection.

One interesting and confounding observation from Figure~\ref{fig:generalization} is how generalization changes for the two problems with boolean program outputs, Compare String Lengths and Mirror Image. Both of these problems, especially Compare String Lengths, exhibited significantly better generalization with novelty-lexicase selection compared to lexicase selection. Let's consider how the novelty scores work for a boolean output problem. Since such a program has only two possible outputs, all of the novelty scores for each test case will have one of two values; whether a novelty score is good or not will depend entirely on whether the output is the same as the majority of the population or the minority of the population.

Now consider how the error case and novelty score case allow an individual to continue or not in the candidates pool during the lexicase process. If the output is correct and is in the minority of outputs in the population, the individual will continue in candidates when either case appears in lexicase. If the output is correct but in the majority, the individual will continue in candidates when the error value appears, but not the novelty score case. On the other hand, if the output is wrong but in the minority, the individual will continue in candidates when the novelty score appears, but not the error case. Thus, being wrong and in the minority is just as good as being right and in the majority, if not better, since the case that allows continuation will have fewer other candidates also continuing! This is quite confounding, as it seems like it rewards being wrong as long as most other individuals are right. It is not clear if this observation contributes to better generalization, and if so, how.



Novelty-lexicase selection not only continually emphasizes unseen parts of the search space, but also targets regions that show promise for solutions, combining the core strengths of both novelty search and lexicase selection. It additionally improves generalization of the solutions it does find. By taking the best parts of novelty search and lexicase selection, novelty-lexicase selection considers more of the search space than lexicase selection but concentrates on objectives more than novelty search, leading to better problem-solving performance than either.

\section{Conclusions and Future Work}
In this paper, we synthesized novelty search and lexicase selection in order to create a new parent selection method, novelty-lexicase selection. In doing so, we hoped to combine the diverging and exploratory qualities of novelty search with fitness seeking of lexicase selection in order to target multiple promising parts of the search space. For novelty-lexicase selection, we calculate the novelty score of every output of every program by simply counting the number of other programs with the same output, instead of calculating the difference between error vectors in the style of novelty search. This makes novelty-lexicase selection less computationally expensive for output types that require significant computation for finding distances, such as strings, making it feasible to run on problems with any output type. To select a parent, novelty-lexicase combines the lists of novelty scores and fitness errors into one shuffled list before applying the lexicase algorithm. Since the lexicase algorithm applies more pressure toward the cases appearing earlier in the shuffled list, novelty-lexicase distributes selection pressure to both individuals with excellent errors and those with novel behaviors, most often selecting individuals that perform well on a combination of the two. This pressure drives evolution to seek novel behavior in productive parts of the search space.

We tested the performance of novelty-lexicase selection by comparing its success rate on ten program synthesis benchmark problems against lexicase selection, tournament selection, and novelty search. We compared the performance of these selection methods using PushGP for a maximum of 300 generations, and found that lexicase selection and novelty-lexicase selection were competitive in solution discovery, with novelty search and tournament selection performing poorly in comparison; additionally, novelty-lexicase selection improved population diversity compared to lexicase selection. We were unable to run novelty search on problems with vector outputs, and using the most informative string distance calculations via Levenshtein distance proved prohibitively slow. Novelty search with Hamming distances between string outputs proved ineffectual. We continued our experimentation by running lexicase selection and novelty-lexicase selection for 1000 generations, where novelty-lexicase selection found significantly more solutions on several of the problems and showed much better generalization rates.

To continue this research, we could imagine changing the novelty-lexicase implementation to more closely follow the traditional novelty calculation. In modifying novelty search to be compatible with lexicase selection, we deviated from the $k$-closest neighbor novelty scoring method. In our novelty score calculation, we simply compare an individual's output on a test case to each other individual's output on the same test case, and check for equality. To give a finer-grained measure of distance to other individuals, we could instead calculate the  novelty error as the \textit{distance} from the individual's output on a test case to the $k$ closest outputs on the same test case. We could imagine this more detailed version of novelty distance could provide a more accurate representation of the novelty of each output, leading to better exploration of novel programs. However, these real-valued distances provide a new challenge for the lexicase algorithm, which performs best if some or many individuals have tied best values on a test case. With these novelty errors, it would likely therefore be important to use $\epsilon$-lexicase selection, which has been shown to handle real-valued errors much better than standard lexicase selection~\cite{LaCava:2018:ecj,  LaCava:2016:GECCO}.

In order to better understand the mechanics of novelty-lexicase selection, we would find it fascinating to track the number of individuals that make it past each step of the lexicase algorithm. In particular, we could compare the number of individuals knocked out of contention by novelty scores compared to fitness errors. It is possible that many times, one or a small number of individuals achieve the best novelty score on each test case; on the other hand, it is just as possible that many individuals have equally good novelty on each score, leading to few individuals being knocked out by novelty scores. A better understanding of the influence of each type of test case on selection could lead to an improved weighted shuffling~\cite{Troise:2017:GPTP} of the novelty scores and fitness errors, instead of simply using uniform shuffling.

In this experiment we used a combination of novelty scores and fitness errors in the list of test cases for the lexicase algorithm, with the goal of simultaneously driving the population toward fit and novel solutions. This combination of novelty and errors improved the novelty-lexicase selection success rate on the problems in our experiments, but could potentially hinder solution discovery for problems with more deceptive (and constrained) fitness landscapes, such as maze problems. Using only novelty scores without fitness errors in lexicase selection could potentially improve upon the performance of novelty search on these types of problems.


\begin{acks}
We would like to thank Alexander Dennis and Lee Spector for conversations that helped shape this work, and to Steven Young for systems support.
\end{acks}

\bibliographystyle{ACM-Reference-Format}
\bibliography{nov-lex} 

\end{document}